# Glyph-aware Embedding of Chinese Characters

**Falcon Z. Dai**[*] and **Zheng Cai**[*]
Toyota Technological Institute at Chicago
dai@ttic.edu, jontsai@ttic.edu

## Abstract

Given the advantage and recent success of English character-level and subword-unit models in several NLP tasks, we consider the equivalent modeling problem for Chinese. Chinese script is logographic and many Chinese logograms are composed of common substructures that provide semantic, phonetic and syntactic hints. In this work, we propose to explicitly incorporate the visual appearance of a character's glyph in its representation, resulting in a novel glyph-aware embedding of Chinese characters. Being inspired by the success of convolutional neural networks in computer vision, we use them to incorporate the spatio-structural patterns of Chinese glyphs as rendered in raw pixels. In the context of two basic Chinese NLP tasks of language modeling and word segmentation, the model learns to represent each character's task-relevant semantic and syntactic information in the character-level embedding.

## 1 Introduction

Recently, in combination with deep learning, character-level and subword-unit-level models has achieved the state-of-the-art performance in various natural language processing (NLP) tasks involving Western languages (Wu et al., 2016), we consider the equivalent modeling problem for solving NLP tasks in Chinese. Unlike English script which is *alphabetic* with a small alphabet, Chinese script is *logographic* with a large set of characters which are meaningful individually. According to *Table of General Standard Characters* (通用规范汉字表) compiled by the Chinese government in 2013, there are 3,500 level-1 (being the most common) characters and more than 8,105 characters in total (Wikipedia, 2017). At the same time, it is not correct to treat Chinese characters as equivalent to English words because the distribution of Chinese characters deviate markedly from Zipf's law (Zipf, 1935; Shtrikman, 1994). Furthermore, there is evidence suggesting that segmented Chinese words, - some of them are unigrams -, distribute according to Zipf's law (Xiao, 2008). Arguably, the closest equivalent linguistic unit in English corresponding to a Chinese character is a subword unit, i.e., word fragments.

Furthermore, there is a strong case for modeling at character-level for task involving Chinese corpora, since Chinese text is usually written without word boundaries to indicate the *segmentation* of characters into words. As a consequence, word-segmented corpora is rare. Traditionally, systems are designed to process words as input, so often, a separately trained or hand crafted routine would first segment the contiguous sequence of characters into words as part of the preprocessing. However, this pipeline design might unnecessarily accumulate error due to segmentation ambiguity that can be resolved in a later stage. The trend of *end-to-end* training of differentiable, neural network-based models also enables training character-level models jointly with the rest of the system under the task objective. It is well-known that many Chinese characters' written form, their *glyphs*, share common sub-structures and some of these sub-structure are informative of the semantics, syntactic role and phonetics of the characters. For example, for semantics, 雨 (rain) 雪 (snow) 雹 (hail) 雷 (thunder) all have a sub-structure 雨, which commonly denote meteorological phenomena.[1] For

---

[*] These authors contributed equally and their names are randomly ordered.

[1] A sub-structure such as 雨 in 雪 is called a *radical*.

syntactic roles, 打 (hit) 提 (lift) 抓 (grab) all contain 扌 which is indicative of a verb. For phonetics, 乙 (yǐ) 亿 (yì) 忆 (yì) all share 乙. However, as far as we are aware of, at the time of our work[2], there is no study that explicitly exploits the *spatio-structural* information of a Chinese character's glyph for NLP tasks.[3] In this work, we explore the effect of incorporating glyphs as additional features in the context of two common Chinese NLP tasks, segmentation and language modeling, resulting in a novel *glyph-aware embedding* of Chinese characters. This work's major contributions are

- a novel character embedding model that explicitly incorporates visual appearance of Chinese characters.

- new state-of-the-art results on a segmentation benchmark task.

## 2 Hypotheses

We hypothesize that the semantic and syntactic information of sub-glyph structures can help improve the character embeddings and thus improve performance in Chinese NLP tasks.

Intuitively, representing each character only by their ID's implies that any pair of characters are as distinct as any other pair. This ignores any common sub-glyph structures shared by characters. Therefore incorporating the glyph's visual information we should be able to generalize knowledge learned about a character to another via their shared sub-glyph structures.

However, this hypothesis is not trivial because there are many Chinese characters that share strikingly similar visual appearances yet not their meanings. For example, 土 (soil) ↔ 士 (roughly means -er as fighter translates to 斗 (fight) 士), and 人 (person) ↔ 入 (enter). By identifying a character with only its visual appearance, we are vulnerable to this new source of ambiguity which can harm performance. Due to this concern, we also include a mixed embedding in our experiments which combine both ID and glyph representation.

---

[2]Since then, we discovered two independent, concurrent studies with approaches similar to ours by Liu et al. (2017) and Costa-jussà et al. (2017).

[3]A character's visual appearance is essential in solving hand-writing recognition tasks which are challenges in computer vision.

## 3 Method

In keeping with the common neural network model architectures, we decided to feed the glyph as an input to a feed-forward neural network (FNN) model, an *embedder*, that outputs an *embedding vector* which, in both the segmentation task and the language modeling task, is then consumed by a recurrent neural network to make predictions. In order to compare the proposed glyph-aware embeddings with the glyph-unaware embeddings, we shall keep the recurrent neural network (RNN) architecture fixed and only change the embedder in our experiments.

Considering that there are many different layouts for sub-glyph structures[4], and the same radical can appear at different positions[5], we think the most promising representation that preserves both the *identities* and the *spatial arrangement* of sub-structures is to use the raw pixels of a glyph.

Being inspired by the success of *convolutional neural networks* (CNN) (LeCun et al., 1995) in learning feature representation in computer vision (Krizhevsky et al., 2012), we used CNN to implement the embedder (see Figure 1). We believe that the spatial translational invariance induced by CNN's filter structure is particularly suited for modeling radicals that can appear at different locations of a glyph. After the CNN, a fully connected layer outputs an embedding vector of some dimension $k$. To apply our method, we first render the glyph for a character using a font file[6] and then feed the glyph as a gray-scale image into the CNN embedder.

We implemented our models and experiments efficiently with Tensorflow (Abadi et al., 2016). In particular, we cached rendered glyphs to reduce repeated render calls of the same character by 1,000 times. We open-source our implementation[7] for replicability

## 4 Results

**Chinese language modeling**

Following the common approach in language modeling (LM), we model the likelihood of a sentence

---

[4]昌 has a vertical layout, 明, horizontal, and 晶, compound.

[5]the radical 口 (mouth) can appear on the left 喊, top 员, bottom 含, inner 向.

[6]We used Google's free Noto font (Google Inc.) throughout this work including the Chinese characters rendered in this paper.

[7]http://github.com/falcondai/chinese-char-lm

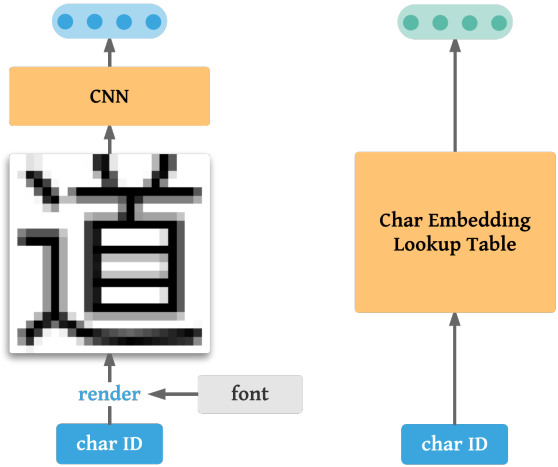

Figure 1: Left: our proposed glyph-aware CNN embedder. Right: the commonly used embedding model (we refer to this as ID embedder). The trainable parameters are labeled in orange.

as

$$p(c_1, \cdots, c_n) = p(c_1) \prod_{i=2}^{n} p(c_i|c_1, \cdots, c_{i-1})$$

where $c_i$ is the $i$-th character in a sentence of $n$ characters. The conditional distribution of $p(c_i|c_1, \cdots, c_{i-1})$ is modeled as a gated recurrent unit (GRU) (Chung et al., 2014) together with an embedder. In all the experiments, we used a GRU with a 128-dimensional hidden state, and 300-dimensional embedding vectors for all embedders. For the CNN embedder, we use a two layer CNN: 32 (7, 7) filters with (2, 2) stride in the first layer, 16 (5, 5) filters (2, 2) stride in the second layer, and a fully-connected layer at the end. For all the layers, we use ReLU non-linearity throughout (Nair and Hinton, 2010). For the linear embedder, we used only one fully-connected layer. For the last row "ID + CNN embedder" in Table. 1, we combine the embedding vectors output by ID and CNN embedders via vector addition. In all the runs, we limited the vocabulary size to 4000 with one unknown class.

We experimented with language modeling on the Microsoft Research dataset (MSR) from the Second International Chinese Word Segmentation Bakeoff (Emerson, 2005). First, we should note that the CNN embedder outperformed the linear embedder by a large margin (see the second and the third row in Table. 1. This is expected as the CNN is more suitable for modeling image data. Second, the ID embedder (see the first row in Table. 1) remains a very strong baseline and the mixed embedder is only as good as the ID embedder by itself (see the fourth row in Table. 1). It seems that CNN embedder did *not* provide extra information useful for the task.

| embedders | test perplexity |
|---|---|
| ID embedder | **47.53** |
| linear embedder | 71.51 |
| CNN embedder | 55.51 |
| ID + CNN embedder | **47.75** |

Table 1: LM performance of different embedders on the test split of MSR.

**Chinese word segmentation**

We use Peking University dataset (PKU) and Microsoft Research dataset (MSR) from the Second International Chinese Word Segmentation Bakeoff(Emerson, 2005) to compare the proposed CNN embedder with the ID embedder. We formulated the segmentation task as a structured prediction problem of predicting whether to insert word boundary behind a character for each character given the whole input sentence. An example would be:

$$\begin{array}{ccccc} 这 & 是 & 一 & 句 & 话 。\\ 1 & 1 & 0 & 1 & 1 \end{array}$$

We experimented with both single-directional GRU and bidirectional long short-term memory (LSTM) recurrent networks (Graves and Schmidhuber, 2005; Hochreiter and Schmidhuber, 1997; Schuster and Paliwal, 1997) as the sequence prediction models in our experiments (RNN segmentor). (see Table. 2 and Table. 3). RNN segmentor takes sequence of embeddings from embedder. For the CNN embedder, we used a single layer ReLU-gated CNN: 16 (5,5) filters with (2,2) stride and a fully-connected layer to output a 100-dimensional embedding vector at the end. For the RNN segmentor, the hidden unit is set to be 100 dimensional with a fully-connected layer mapping the output hidden state to a binary prediction at each character. Overall, on both PKU and MSR, the proposed mixed embedder and bidirectional LSTM achieved the best performance outperforming the previous state-of-the-art on by a significant margin. Similar to the LM experiments, we use a vocabulary of 4000 and one unknown class.

| RNN segmentors | embedder | precision | recall | F1 |
|---|---|---|---|---|
| GRU | ID | 87.41 | 84.14 | 85.75 |
| | CNN | 90.03 | **89.54** | **89.78** |
| | ID + CNN | **90.46** | 88.80 | 89.62 |
| Bidirectional LSTM | ID | 96.06 | 94.66 | 95.36 |
| | CNN | 94.73 | 94.88 | 94.81 |
| | ID + CNN | **96.91** | **95.41** | **96.15** |
| NWS (Cai and Zhao, 2016) | | 95.5 | 94.9 | 95.16 |

Table 2: segmentation results on PKU dataset

| RNN segmentors | embedder | precision | recall | F1 |
|---|---|---|---|---|
| GRU | ID | 86.97 | 85.25 | 86.10 |
| | CNN | **89.93** | 86.79 | **88.33** |
| | ID + CNN | 88.81 | **87.19** | 88.00 |
| Bidirectional LSTM | ID | 97.34 | **97.25** | 97.29 |
| | CNN | 97.07 | 96.98 | 97.03 |
| | ID + CNN | **97.82** | 97.04 | **97.43** |
| NWS (Cai and Zhao, 2016) | | 96.1 | 96.7 | 96.4 |

Table 3: segmentation results on MSR dataset

We use Adam (Kingma and Ba, 2014) optimizer throughout all our experiments.

## 5 Analysis

Due to the lack of improvement of the proposed mixed embedder over the ID embedder in the language modeling task, we suspect that the CNN embedder is under-trained. Unlike a digit class in MNIST (LeCun et al., 2010) which has 6,000 training examples, given one font, a character only has one glyph and every sub-glyph structure appears on average in only about 40 characters. Thus we suspect that the variability in input to the CNN is too limited. Modeling after common image augmentation technique (Krizhevsky et al., 2012), we applied random jitters, i.e., 2D translation with $\Delta x, \Delta y \in \{-2, -1, 0, +1, +2\}$, to the input glyphs at training time. This increases the input variations by 25-fold but the perplexity degrades slightly to 49.66.

Since we mix the ID embedding and CNN embedding by summation in the proposed mixed embedder, the norm of each component embedding determines the relative importance of that representation in the resulting embedding. In Figure 2, we observe that the CNN embeddings distribute differently in the trained segmentation model and the trained language modeling model. In the case of language modeling, the norm of CNN embeddings is squashed suggesting that CNN embedding is largely ignored.

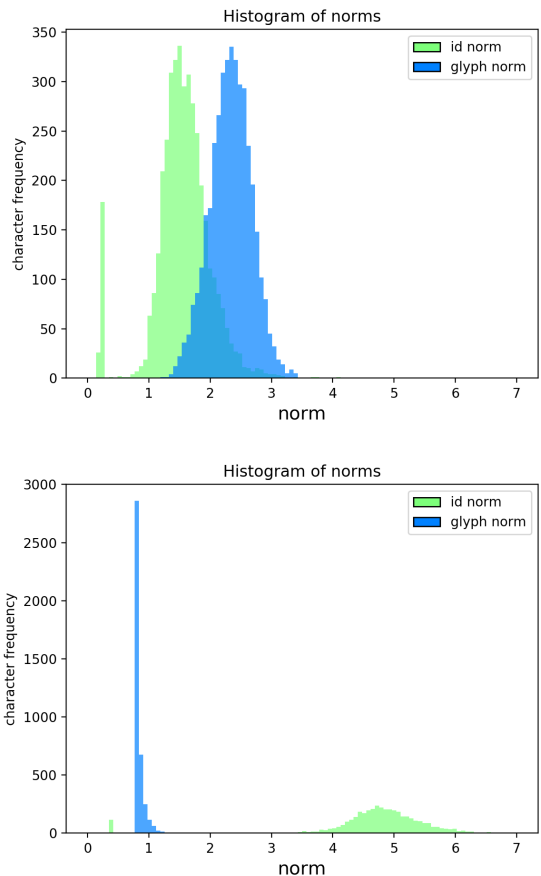

Figure 2: The distribution of the Frobenius norm of ID embeddings (id norm) and CNN embeddings (glyph norm) from the mixed embedder. Top: the segmentation task. Bottom: the language modeling task.

## 6 Discussion

It should be noted that the number of parameters of the proposed CNN embedder is different than that of the ID embedder. Suppose the dimensionality of the embedding vectors is $K$, and the vocabulary size is $N$, the CNN embedder has $O(N + K)$ many parameters: $O(K)$ many *trainable* parameters and $O(N)$ glyphs rendered from a font file. In contrast, the ID embedder has $O(NK)$ many parameters, all of which are *trainable*. This means that the CNN embedder is a more compact representation with competitive performance as the ID embedder.

**Related work**

Shi et al. (2015) represented a character by its radicals based on Wubi input method but this ignores the scales and spatial arrangement of each radical which are present in our rendered glyphs.

It came to our late attention that independently, Liu et al. (2017) considered the same character-level modeling problem and experimented with vanilla CNN models almost identical to ours. They evaluated their method on a new document classification task instead of the commonly considered tasks or benchmarks we considered in this work. Consistent with their findings, we also observed similar effects of CNN embedder, ID embedder and mixed embedder in our tasks. Our mixed embedded corresponds roughly to their early fusion model. Costa-jussà et al. (2017) also considered incorporating Chinese glyphs as additional features in their Chinese-Spanish machine translation system and their modeling approach corresponds roughly to our linear embedder.

**Future work**

We hope to delve deeper into the cause of the CNN embedder's low performance in the LM task. In particular, we want to experiment with using bag-of-stroke prediction in a multi-task loss to provide CNN with extra supervision during training. Furthermore, we have only explored two NLP tasks that emphasize semantic and syntactic information in this work. In the future, we hope to explore tasks that requires more phonetic information to do well, such as phoneme prediction.

## 7 Conclusion

Our experiments show that glyph-aware embedding can improve performance in some Chinese NLP tasks, in particular, the word segmentation task. Further studies are needed to understand the usefulness of glyph features in a more comprehensive way. However, given the visual ambiguity inherent in Chinese characters and the difficulty to interpret neural network models, any further research that uses glyph features and deep learning methods should exercise caution when measuring and verifying the contribution of the glyph features.

**Acknowledgement**

We thank Kevin Gimpel, Matthew Walter and Karen Livescu for their kind support and helpful comments during our investigation. We also thank the anonymous reviewers for their constructive suggestions.

**References**

Martín Abadi, Ashish Agarwal, Paul Barham, Eugene Brevdo, Zhifeng Chen, Craig Citro, Greg S. Corrado, Andy Davis, Jeffrey Dean, Matthieu Devin, and others. 2016. Tensorflow: Large-scale machine learning on heterogeneous distributed systems. *CoRR*.

Deng Cai and Hai Zhao. 2016. Neural word segmentation learning for chinese. *CoRR*.

Junyoung Chung, Caglar Gulcehre, KyungHyun Cho, and Yoshua Bengio. 2014. Empirical evaluation of gated recurrent neural networks on sequence modeling. *CoRR*.

Marta R Costa-jussà, David Aldón, and José AR Fonollosa. 2017. Chinese–spanish neural machine translation enhanced with character and word bitmap fonts. *Machine Translation*, pages 1–13.

Tom Emerson. 2005. Second international chinese word segmentation bakeoff. http://sighan.cs.uchicago.edu/bakeoff2005/. Accessed: 2017-04-15.

Google Inc. Google noto fonts. https://www.google.com/get/noto/help/cjk/. Accessed: 2017-07-21.

Alex Graves and Jürgen Schmidhuber. 2005. Framewise phoneme classification with bidirectional lstm and other neural network architectures. *Neural Networks*, 18(5):602–610.

Sepp Hochreiter and Jürgen Schmidhuber. 1997. Long short-term memory. *Neural computation*, 9(8):1735–1780.

Diederik Kingma and Jimmy Ba. 2014. Adam: A method for stochastic optimization. *CoRR*.

Alex Krizhevsky, Ilya Sutskever, and Geoffrey E Hinton. 2012. ImageNet Classification with Deep Convolutional Neural Networks. In F. Pereira, C. J. C. Burges, L. Bottou, and K. Q. Weinberger, editors, *Advances in Neural Information Processing Systems 25*, pages 1097–1105. Curran Associates, Inc.

Yann LeCun, Yoshua Bengio, et al. 1995. Convolutional networks for images, speech, and time series. *The handbook of brain theory and neural networks*, 3361(10):1995.

Yann LeCun, Corinna Cortes, and Christopher JC Burges. 2010. Mnist handwritten digit database. *AT&T Labs [Online]. Available: http://yann.lecun.com/exdb/mnist*, 2.

Frederick Liu, Han Lu, Chieh Lo, and Graham Neubig. 2017. Learning Character-level Compositionality with Visual Features. *CoRR*. ArXiv: 1704.04859.

Vinod Nair and Geoffrey E Hinton. 2010. Rectified linear units improve restricted boltzmann machines. In *Proceedings of the 27th international conference on machine learning (ICML-10)*, pages 807–814.

Mike Schuster and Kuldip K Paliwal. 1997. Bidirectional recurrent neural networks. *IEEE Transactions on Signal Processing*, 45(11):2673–2681.

Xinlei Shi, Junjie Zhai, Xudong Yang, Zehua Xie, and Chao Liu. 2015. Radical embedding: Delving deeper to chinese radicals. In *Proceedings of the 53rd Annual Meeting of the Association for Computational Linguistics and the 7th International Joint Conference on Natural Language Processing (Volume 2: Short Papers)*, pages 594–598, Beijing, China. Association for Computational Linguistics.

S. Shtrikman. 1994. Some comments on Zipf's law for the Chinese language. *Journal of Information Science*, 20(2):142–143.

Wikipedia. 2017. Table of general standard chinese characters — wikipedia, the free encyclopedia. [Online; accessed 21-July-2017].

Yonghui Wu, Mike Schuster, Zhifeng Chen, Quoc V. Le, Mohammad Norouzi, Wolfgang Macherey, Maxim Krikun, Yuan Cao, Qin Gao, Klaus Macherey, Jeff Klingner, Apurva Shah, Melvin Johnson, Xiaobing Liu, Łukasz Kaiser, Stephan Gouws, Yoshikiyo Kato, Taku Kudo, Hideto Kazawa, Keith Stevens, George Kurian, Nishant Patil, Wei Wang, Cliff Young, Jason Smith, Jason Riesa, Alex Rudnick, Oriol Vinyals, Greg Corrado, Macduff Hughes, and Jeffrey Dean. 2016. Google's neural machine translation system: Bridging the gap between human and machine translation. *CoRR*, abs/1609.08144.

Hang Xiao. 2008. On the Applicability of Zipf's Law in Chinese Word Frequency Distribution. *Journal of Chinese Language and Computing*, 18(1):33–46.

George K Zipf. 1935. The psychology of language. *NY Houghton-Mifflin*.